\documentclass{article}

\usepackage{microtype}
\usepackage{graphicx}
\usepackage{subfigure}
\usepackage{booktabs} 

\usepackage{hyperref}

\usepackage[accepted]{icml2023}

\usepackage{amsmath}
\usepackage{amssymb}
\usepackage{mathtools}
\usepackage{amsthm}

\usepackage[capitalize,noabbrev]{cleveref}

\theoremstyle{plain}

\theoremstyle{definition}

\theoremstyle{remark}

\usepackage[textsize=tiny]{todonotes}

\usepackage{xspace}
\usepackage{multirow}

\newcommand{\nt}{\nabla_\theta}
\newcommand{\E}{\mathop{\mathbb{E}}}
\newcommand{\ptf}{p_t^\mathrm{f}}
\newcommand{\pTf}{p_T^\mathrm{f}}
\newcommand{\ptpf}{p_t^\mathrm{pf}}

\newcommand{\npe}{NPE\xspace}
\newcommand{\nle}{NLE\xspace}
\newcommand{\nre}{NRE\xspace}
\newcommand{\nspe}{NPSE\xspace}
\newcommand{\fnspe}{F-NPSE\xspace}
\newcommand{\pfnspe}{PF-NPSE\xspace}

\DeclarePairedDelimiter\ceil{\lceil}{\rceil}

\icmltitlerunning{Compositional Score Modeling for Simulation-based Inference}

\begin{document}

\twocolumn[
\icmltitle{Compositional Score Modeling for Simulation-Based Inference}



\icmlsetsymbol{equal}{*}

\begin{icmlauthorlist}
\icmlauthor{Tomas Geffner}{intern,umass}
\icmlauthor{George Papamakarios}{dm}
\icmlauthor{Andriy Mnih}{dm}
\end{icmlauthorlist}

\icmlaffiliation{dm}{DeepMind}
\icmlaffiliation{intern}{Work done during an internship at DeepMind.}
\icmlaffiliation{umass}{University of Massachusetts, Amherst.}

\icmlcorrespondingauthor{Tomas Geffner}{tgeffner@cs.umass.edu}
\icmlcorrespondingauthor{Andriy Mnih}{andriy@deepmind.com}

\icmlkeywords{Machine Learning, ICML, Inference}

\vskip 0.3in
]



\printAffiliationsAndNotice{} 

\begin{abstract}
Neural Posterior Estimation methods for simulation-based inference can be ill-suited for dealing with posterior distributions obtained by conditioning on multiple observations, as they tend to require a large number of simulator calls to learn accurate approximations. In contrast, Neural Likelihood Estimation methods can handle multiple observations at inference time after learning from individual observations, but they rely on standard inference methods, such as MCMC or variational inference, which come with certain performance drawbacks. We introduce a new method based on conditional score modeling that enjoys the benefits of both approaches. We model the scores of the (diffused) posterior distributions induced by individual observations, and introduce a way of combining the learned scores to approximately sample from the target posterior distribution. Our approach is sample-efficient, can naturally aggregate multiple observations at inference time, and avoids the drawbacks of standard inference methods.
\end{abstract}

\section{Introduction} \label{sec:intro}

Mechanistic simulators have been developed in a wide range of scientific domains to model complex phenomena \cite{cranmer2020frontier}. Often, these simulators act as a black box: they are controlled by parameters $\theta$ and can be simulated to produce a synthetic observation $x$. Typically, the parameters $\theta$ need to be inferred from data. In this paper, we consider the Bayesian formulation of this problem: given a prior $p(\theta)$ and a set of i.i.d.\ observations $x^o_1, \hdots, x^o_n$, the goal is to approximate the posterior distribution $p(\theta \vert x^o_1, \hdots, x^o_n) \propto p(\theta)\prod_{i=1}^n p(x^o_i|\theta)$.

Simulators can be easily sampled from, but the distribution over the outputs---the likelihood $p(x\vert \theta)$---cannot be generally evaluated, as it is implicitly defined. This renders standard inference algorithms that rely on likelihood evaluations, such as Markov chain Monte Carlo (MCMC) or variational inference, inapplicable. Instead, a family of inference methods that rely solely on simulations, known as simulation-based inference (SBI), have been developed for performing inference with these models \cite{beaumont2019approximate}.

Approximate Bayesian Computation is a traditional SBI method \cite{sisson2018handbook}. Its simplest form is based on rejection sampling, while more advanced variants involve adaptations of MCMC \cite{marjoram2003markov} and sequential Monte Carlo \cite{sisson2007sequential, del2012adaptive}. While popular, these methods often require many simulator calls to yield accurate approximations, which may be problematic with expensive simulators. Thus, recent work has focused on developing algorithms that yield good approximations using a limited budget of simulator calls.

Neural Posterior Estimation (\npe) is a promising alternative \cite{papamakarios2016fast, lueckmann2017flexible, chan2018likelihood}. \npe methods train a conditional density estimator $q(\theta \vert x_1, \hdots, x_n)$ to approximate the target posterior,
using a dataset built by sampling the prior $p(\theta)$ and the simulator $p(x\vert \theta)$ multiple times. These methods have shown good performance when approximating posteriors $p(\theta \vert x^o)$ conditioned on a single observation $x^o$ (i.e.\ $n=1$) \cite{lueckmann2021benchmarking}. However, their efficiency decreases for multiple observations ($n>1$), or when $n$ is not known a priori, as in such cases the simulator needs to be called several times per setting of parameters $\theta$ to generate each training case in the dataset, which is inefficient. 

Neural Likelihood Estimation (\nle) \cite{wood2010statistical, papamakarios2019sequential, lueckmann2019likelihood} is a natural alternative when the goal is to approximate posteriors conditioned on multiple observations. \nle methods train a surrogate likelihood $q(x\vert \theta)$
using samples from $p(x | \theta)$. Then, given observations $x_1^o, \hdots, x_n^o$, inference is carried out using the surrogate likelihood by standard methods, typically MCMC \cite{papamakarios2019sequential, lueckmann2019likelihood} or variational inference \cite{wiqvist2021sequential, glockler2022variational}. In contrast to \npe, \nle methods require a single call to the simulator per training case, and can naturally handle an arbitrary number of i.i.d.\ observations at inference time. However, their performance is hampered by their reliance on the underlying inference method, which can introduce additional approximation error and extra failure modes, such as struggling with multimodal distributions.

In this work, we aim to develop a method that enjoys most of the benefits of existing approaches while avoiding their drawbacks. That is, we aim for a method that (i) can aggregate an arbitrary number of observations at inference time while requiring a low number of simulator calls per training case; and (ii) avoids the limitations of standard inference methods. To this end, we make use of score modeling (also known as diffusion modeling), which has recently emerged as a powerful approach for modeling distributions \cite{sohl2015deep, ho2020denoising, song2019generative}. Score-based methods diffuse the target distribution with Gaussian kernels of increasing noise levels, train a score network to model the score (gradient of the log-density) of the resulting densities, and use the trained network to approximately sample the target. They have shown impressive performance, particularly in text-conditional image generation \cite{nichol2022glide, ramesh2022hierarchical}.

In principle, one could directly apply conditional score-based methods (i.e., conditional diffusion models) \cite{ho2020denoising, song2020score, batzolis2021conditional} to the SBI task, which leads to an approach we call Neural Posterior Score Estimation (\nspe).\footnote{We use this name for consistency with concurrent work by \citet{sharrock2022sequential} that also explores this idea.} However, as explained in \cref{sec:diff_npe}, this fails to satisfy our desiderata. We address this by proposing a different destructive/forward process to the one typically used by diffusion models. Simply put, we factorize the distribution $p(\theta\vert x_1, \hdots, x_n)$ in terms of the posterior distributions induced by  individual observations $p(\theta \vert x_i)$, train a conditional score network to approximate the score of (diffused versions of) $p(\theta \vert x)$ for any single $x$, and propose an algorithm that uses the trained network to approximately sample from the posterior $p(\theta\vert x^o_1, \hdots, x^o_n)$ for \textit{any} number of observations $n$. Our method satisfies our desiderata: it can naturally handle sets of observations of arbitrary sizes without increasing the simulation cost, and it avoids the limitations of standard inference methods by using an annealing-style sampling algorithm \cite{sohl2015deep, song2019generative, ho2020denoising}. We describe our approach, called Factorized Neural Posterior Score Estimation (\fnspe), in detail in \cref{sec:ndpe}.

Additionally, a simple analysis suggests that \nspe and \fnspe should not be seen as independent, but as the two extremes of a spectrum of methods. We present this analysis in \cref{sec:pndpe}, where we also propose a family of approaches, called Partially Factorized Neural Posterior Score Estimation (\pfnspe), that populates this spectrum. 

Finally, \cref{sec:exps} presents a comprehensive empirical evaluation of the proposed approaches on a range of tasks typically used to evaluate SBI methods \cite{lueckmann2021benchmarking}. Our results show that our proposed methods tend to outperform relevant baselines when multiple observations are available at inference time, and that the use of methods in the \pfnspe family often leads to increased robustness.

\section{Preliminaries} \label{sec:prels}

\subsection{Simulation-based Inference} \label{sec:prels_sbi}

\paragraph{Neural Posterior Estimation (NPE).}

\npe methods use a conditional neural density estimator, typically a normalizing flow \cite{tabak2013family, rezende2015variational, winkler2019learning}, to approximate the target posterior. When observed data consists of a single sample $x^o$, the $\psi$-parameterized density estimator $q_\psi(\theta \vert x)$ is trained via maximum likelihood, maximizing $\E_{p(\theta) p(x\vert \theta)} \log q_\psi(\theta \vert x)$ with respect to $\psi$. Since this expectation is intractable, \npe replaces it with an empirical approximation over a dataset $\{\theta^i, x^i\}_{i=1}^M$, where $(\theta^i, x^i) \sim p(\theta) p(x\vert \theta)$. Since the conditional neural density estimator takes as input both $\theta$ and $x$, models $q_\psi(\theta\vert x)$ trained this way provide an amortized approximation to the target posterior distribution: at inference time, $q_\psi(\theta\vert x^o)$ yields an approximation of $p(\theta \vert x^o)$ for \textit{any} observation $x^o$.

The situation changes when we have $n>1$ observations at inference time, since it is often not clear how to combine approximations $q_\psi(\theta \vert x^o_i) \approx p(\theta \vert x^o_i)$ to get a tractable approximation of $p(\theta \vert x^o_1,\ldots,x^o_n)$. Instead, the density estimator has to take a set of $n$ observations as conditioning, $q_\psi(\theta \vert x_1, \ldots, x_n)$, and training has to be done on samples $(\theta^i, x_1^i, \ldots, x_n^i) \sim p(\theta) \prod_j p(x_j\vert \theta)$ \cite{chan2018likelihood}. While the learned density estimator provides an amortized approximation to the target posterior, this comes at the price of reduced sample efficiency, as generating each training case requires $n$ simulator calls per parameter setting $\theta$.

\npe methods can also be used when the number of observations $n$ is not known a priori \cite{radev2020bayesflow}. In such cases the density estimator is trained to handle from $n=1$ to $n_{\max}$ observations. This can be achieved by parameterizing $q_\psi(\theta \vert x_1, \ldots, x_n) = q_\psi(\theta \vert h_\psi(x_1, \ldots, x_n), n)$ with a permutation-invariant function $h_\psi$ \cite{zaheer2017deep}, and learning from training cases with a variable number of observations $(n^i, \theta^i, x_1^i, \ldots, x^i_{n^i}) \sim \mathcal{U}(n;n_{\max}) p(\theta) \prod_{j=1}^n p(x_j \vert \theta)$, where $\mathcal{U}(\cdot ;n_{\max})$ is the uniform distribution over $\{1, 2, \ldots, n_{\max}\}$. Then, at inference time, the density estimator provides an approximation of $p(\theta \vert x^o_1, \ldots, x^o_{n_o})$ for any set of observations $x^o_1, \ldots, x^o_{n_o}$ with cardinality $n_o \in \{1, 2, \ldots, n_{\max}\}$. This approach requires, on average, $n_{\max}/2$ simulator calls per training case.

\paragraph{Neural Likelihood Estimation (NLE).}

Instead of approximating the posterior distribution directly, \nle methods train a surrogate $q_\psi(x \vert \theta)$ for the likelihood $p(x \vert \theta)$. The surrogate is trained with maximum likelihood on samples $(\theta^i, x^i) \sim \tilde{p}(\theta) p(x\vert \theta)$, where $ \tilde{p}(\theta)$ is a proposal distribution with sufficient coverage, which can in the simplest case default to the prior. Then, given an arbitrary set of i.i.d.\ observations $x_1^o, \ldots, x_n^o$, inference is carried out by running MCMC or variational inference on the approximate target
\begin{equation} \label{eq:post_fact_nle}
    p(\theta) \prod_{i=1}^n q_\psi(x_i^o \vert \theta),
\end{equation}
obtained by replacing the individual likelihoods $p( x_i^o \vert \theta)$ in the exact posterior expression $p(\theta \vert x_1^o, \ldots, x_n^o) \propto p(\theta) \prod_{i=1}^n p(x_i^o\vert \theta)$ by the learned approximation $q_\psi(x_i^o \vert \theta)$.

A key benefit of \nle is the ability to aggregate multiple observations at inference time, while only training on single-observation/parameter pairs. This is achieved by exploiting the posterior's factorization in terms of the individual likelihoods in \cref{eq:post_fact_nle}. However, the reliance on MCMC or variational inference can negatively impact \nle's performance, as it introduces additional approximation error and potential failure modes. For instance, a failure mode often reported for \nle \citep[e.g.\ by][]{greenberg2019automatic} involves its inability to robustly handle multimodality (we also observe this in our empirical evaluation in \cref{sec:exps}).

\paragraph{Neural Ratio Estimation (NRE).} Prior work has proposed to learn likelihood ratios \cite{pham2014note, cranmer2015approximating} instead of the likelihood, and to use the learned ratios to perform inference. This approach retains \nle's ability to aggregate multiple observations at inference time while training on single-observation/parameter pairs, and may be more convenient than \nle when learning the full likelihood is hard, e.g.\ when observations are high-dimensional. However, NRE methods still rely on standard inference techniques, which can hurt their performance.

\subsection{Conditional Score-based Generative Modeling} \label{sec:sbgm}

This section introduces conditional score-based methods for generative modeling, the main tool behind our approach. The goal of conditional generative modeling is to learn an approximation to a distribution $p(\theta \vert c)$, for some conditioning variable $c$, given samples $(\theta, c) \sim p(\theta, c)$, which is the problem SBI methods need to solve. Methods based on score modeling have shown impressive performance for this task \cite{dhariwal2021diffusion, ho2022classifier, ramesh2022hierarchical, saharia2022photorealistic}. They define a sequence of densities $p_0(\theta\vert c), \hdots, p_T(\theta \vert c)$ by diffusing the target $p(\theta \vert c)$ with increasing levels of Gaussian noise, learn the scores of each density in the sequence via denoising score matching \cite{hyvarinen2005estimation, vincent2011connection}, and use variants of annealed Langevin dynamics \cite{roberts1996langevin, welling2011bayesian} with the learned scores to approximately sample from the target distribution.

Specifically, given $0 \approx \gamma_T < \gamma_{T-1} < \cdots < \gamma_1 < 1$ and the corresponding Gaussian diffusion kernels $p_t(\theta\vert \theta') = \mathcal{N}(\theta \vert \sqrt{\gamma_t} \, \theta', (1 - \gamma_t)I)$, 
the sequence of densities used by score-based methods is typically defined as
\begin{equation} \label{eq:sbmbr}
\begin{split}
\textstyle p_0(\theta \vert c) & = p(\theta \vert c) \\
p_t(\theta \vert c) & = \int d\theta' \, p(\theta'\vert c) p_t(\theta\vert \theta'),
\end{split}
\end{equation}
for $t=1, \ldots, T$. Since $\gamma_T \approx 0$, this sequence can be seen as gradually bridging between the tractable reference $\mathcal{N}(\theta \vert 0, I) \approx p_T(\theta)$ and the target $p(\theta\vert c) = p_0(\theta \vert c)$. Score-based methods then train a score network $s_\psi(\theta, t, c)$ parameterized by $\psi$ to approximate the scores of these densities, $\nabla_\theta \log p_t(\theta\vert c)$, by minimizing the denoising score matching objective \cite{hyvarinen2005estimation, vincent2011connection} 
\begin{equation}
    \sum_{t=1}^{T-1} \E_{p(c, \theta') p_t(\theta \vert \theta')} \left [ \lambda(t) \left \Vert s_\psi(\theta, t, c) - \nabla_{\theta} \log p_t(\theta \vert \theta')\right \Vert^2 \right ],
\end{equation}
where $\lambda(t)$ is some non-negative weighting function \cite{ho2020denoising, song2021maximum}. Finally, the learned score network is used to approximately sample the target using annealed Langevin dynamics, as shown in \cref{alg:sampleuld}. Other, but still related, sampling algorithms can be derived by analysing score-based methods from the perspective of diffusion processes \cite{ho2020denoising, song2020score}.

\begin{algorithm}[ht]
\caption{Annealed Langevin with learned scores}
\label{alg:sampleuld}
\begin{algorithmic}[1]
\STATE {\bfseries Input:} Score network $s_\psi(\theta, t, c)$
\STATE {\bfseries Input:} Reference $p_T(\theta)$, conditioning variable $c$
\STATE {\bfseries Input:} Number of Langevin steps $L$ and step sizes $\delta_t$
\STATE $\theta \sim p_T(\theta)$ \hfill $\triangleright$ Sample reference
\FOR{$t = T-1, T-2, \hdots, 1$}
	\FOR{$s = 1, 2, \hdots, L$}
	    \STATE $\eta_{ts} \sim \mathcal{N}(0, I)$ \hfill $\triangleright$ Sample noise
	    \STATE $\theta \leftarrow \theta + \frac{\delta_t}{2} s_\psi(\theta, t, c) + \sqrt{\delta_t} \eta_{ts}$ \hfill $\triangleright$ Langevin step
    \ENDFOR
\ENDFOR
\end{algorithmic}
\end{algorithm}

\section{Score Modeling for SBI} \label{sec:ndpe}

This section presents \fnspe, our approach to SBI. Our goal is to develop a method that (i) can aggregate an arbitrary number of observations at inference time while requiring a low number of simulator calls per training case; and (ii) avoids the limitations of generic inference methods. We achieve this by building on conditional score modeling, but using a different construction for the bridging densities and reference distribution from the ones typically used. \Cref{sec:ndpe_sub} presents our approach, and \cref{sec:diff_npe} explains why this novel construction is necessary, by showing that the direct application of the score modeling framework to the SBI task (i.e.\ \nspe), fails to satisfy (i). Finally, \cref{sec:pndpe} presents \pfnspe, a family of methods that generalizes \fnspe and \nspe by interpolating between them.

\subsection{Factorized Neural Posterior Score Estimation} \label{sec:ndpe_sub}

\fnspe builds on the conditional score modeling framework, but with a different construction for the bridging densities and reference distribution. Our construction is based on the  factorization of the posterior in terms of the individual observation posteriors (see \cref{app:derfact}):
\begin{equation} \label{eq:postfact}
p(\theta \vert x_{1}, \ldots, x_{n}) \propto p(\theta)^{1-n} \prod_{j=1}^n p(\theta \vert x_{j}).
\end{equation}
The main idea behind our approach is to define bridging densities that satisfy a similar factorization. To achieve this we propose a sequence indexed by $t=0, \ldots, T$ given by
\begin{equation} \label{eq:approachbr}
    \ptf(\theta \vert x_1, \hdots, x_n) \propto \left(p(\theta)^{1-n}\right)^{\frac{T-t}{T}} \prod_{j=1}^n p_t(\theta \vert x_{j}),
\end{equation}
where $p_t(\theta \vert x_{j})$ follows \cref{eq:sbmbr} with $c=x_j$, and the superscript $\mathrm{f}$ is used as an identifier of \fnspe.

This construction has four key properties. First, the distribution for $t=0$ recovers the target $p(\theta \vert x_{1}, \ldots, x_{n})$. Second, the distribution for $t=T$ approximates a spherical Gaussian $\pTf(\theta \vert x_1, \hdots, x_n) \approx p_T(\theta) = \mathcal{N}(\theta\vert 0, \frac{1}{n} I)$, since the prior term vanishes and  $p_T(\theta \vert x_j) \approx \mathcal{N}(\theta\vert 0, I)$, and thus can be used as a tractable reference for the diffusion process. Third, the scores of the resulting densities can be decomposed in terms of the score of the prior (typically available exactly) and the scores of the individual terms $p_t(\theta \vert x_j)$ as 
\begin{multline}
    \nt \log \ptf(\theta \vert x_1, \hdots, x_n) = \\ \frac{(1-n)(T-t)}{T} \nt \log p(\theta) + \sum_{j=1}^n \nt \log p_t(\theta \vert x_{j}).
\end{multline}
And fourth, the scores $\nt \log p_t(\theta \vert x_{j})$ can all be approximated using a \textit{single} score network $s_\psi(\theta, t, x)$ trained via denoising score matching on samples $(\theta^i, x^i) \sim p(\theta) p(x\vert \theta)$, as explained in \cref{sec:sbgm}.

After training, given i.i.d.\ observations $x^o_1, \hdots, x^o_n$, we can approximately sample $p(\theta \vert x^o_1, \hdots, x^o_n)$ by running \cref{alg:sampleuld} with the reference $p_T(\theta) = \mathcal{N}(\theta\vert 0, \frac{1}{n} I)$, conditioning variable $c = \{x^o_1, \hdots, x^o_n\}$, and approximate score
\begin{multline} \label{eq:scoreapprox}
    s_\psi(\theta, t, c) = \\ \frac{(1-n)(T-t)}{T} \nt \log p(\theta) + \sum_{j=1}^n s_\psi(\theta, t, x^o_j).
\end{multline}
In short, the fact that the bridging densities factorize over individual observations as in \cref{eq:approachbr} allows us to train a score network to approximate the scores of the distributions induced by individual observations, and to aggregate the network's output for different observations at inference time to sample from the target posterior distribution. Thus, we can aggregate an arbitrary number of observations at inference time while training on samples $(\theta^i, x^i)\sim p(\theta) p(x\vert \theta)$, each one requiring a single simulator call. Additionally, the mass-covering properties of score-based methods \cite{ho2020denoising, song2021maximum} together with the annealed sampling algorithm avoid the drawbacks of standard inference methods, such as their difficulty with handling multimodality.

As presented, applying \fnspe to models with constrained priors (e.g. Beta, uniform) requires care, as the densities in \cref{eq:approachbr} are ill-defined outside of the prior's support. This can be addressed in two ways: (i) reparameterizing the model such that the prior becomes a standard Gaussian; this is often easy to do, and what we do in this work, see \cref{app:models}; or (ii) diffusing the prior and learning the corresponding scores. Concurrent work by \citet{sharrock2022sequential} explored (ii), obtaining good results.

A potential drawback of \fnspe is that it might accumulate errors when combining score estimates as in \cref{eq:scoreapprox}, affecting the method's performance. This drawback is shared by \nle and \nre. We study it empirically in \cref{sec:exps}. 

\subsection{Direct Application of Conditional Score Modeling} \label{sec:diff_npe}

This section introduces \nspe and explains why it fails to satisfy our desiderata, specifically (i). \nspe is a direct application of the score modeling framework to the SBI task. Following \cref{eq:sbmbr} with $c = \{x_1, \ldots, x_n\}$, and assuming a fixed number of observations $n$ (relaxed later), \nspe defines a sequence of densities
\begin{equation} \label{eq:score_naive}
    p_t(\theta \vert x_1, \hdots, x_n) = \int d\theta' \, p(\theta'\vert x_1, \hdots, x_n) p_t(\theta\vert \theta'),
\end{equation}
and trains a score network $s_\psi(\theta, t, x_1, \ldots, x_n)$ to approximate their scores via denoising score matching. Then, at inference time, given observations $x_1^o,\ldots,x_n^o$, \nspe uses $s_\psi(\theta, t, x^o_1, \ldots, x^o_n)$ to approximately sample $p(\theta \vert x_1^o, \ldots, x_1^o)$. The approach can be extended to cases where $n$ is not fixed a priori, by parameterizing the score network as $s_\psi(\theta, t, x_1, \ldots, x_n) = s_\psi(\theta, t, h_\psi(x_1, \ldots, x_n), n)$ with a permutation-invariant function $h_\psi$, and training it on samples $(n^i, \theta^i, x_1^i, \ldots, x^i_{n^i}) \sim \mathcal{U}(n;n_{\max}) p(\theta) \prod_{j=1}^n p(x_j \vert \theta)$.

Unfortunately, \nspe does not satisfy condition (i) outlined above, as it requires specifying the maximum number of observations $n_{\max}$ and needs $n_{\max} / 2$ simulator calls per training case on average. Since the scores of the bridging densities defined in \cref{eq:score_naive} do not factorize in terms of the individual observations, the approach taken by \fnspe, which trains a score network for individual observations and aggregates them at inference time, is not applicable. However, in contrast to \fnspe, \nspe does not require summing approximations, so it does not accumulate errors.

\subsection{Partially Factorized Neural Posterior Score Estimation} \label{sec:pndpe}

We introduced \nspe and \fnspe, and described their benefits and limitations; \fnspe is efficient in terms of the number of simulator calls but may accumulate approximation error, while the opposite is true for \nspe. We argue that these two approaches can be seen as the opposite extremes of a family of methods that interpolate between them, achieving different trade-offs between sample efficiency and accumulation of errors. We call these methods Partially Factorized Neural Posterior Score Estimation (\pfnspe).

\pfnspe is based on a similar strategy to the one used by \fnspe, factorizing the target posterior distribution in terms of small subsets of observations instead of individual observations. Specifically, given $m\geq 1$, we partition the set of conditioning variables $\{x_1, \ldots, x_n\}$ into $k=\ceil{n/m}$ disjoint subsets of size at most $m$, denoted by $X_{1}, \ldots, X_{k}$, and factorize the posterior distribution as
\begin{equation} \label{eq:partialpostfact}
p(\theta \vert x_{1}, \ldots, x_{n}) \propto p(\theta)^{1-k} \prod_{j=1}^k p(\theta \vert X_{j}).
\end{equation}
Then, following the \fnspe strategy, we define the bridging densities using the diffused versions of each $p(\theta \vert X_{j})$ as 
\begin{equation} \label{eq:parttialapproachbr}
    \ptpf(\theta \vert x_1, \hdots, x_n) \propto \left(p(\theta)^{1-k}\right)^{\frac{T-t}{T}} \prod_{j=1}^k p_t(\theta \vert X_{j}),
\end{equation}
and train a score network $s_\psi(\theta, t, X_j)$ to approximate their scores. Since this network needs to handle  input sets $X_j$ of sizes varying between $1$ and $m$, we parameterize it using a permutation-invariant function, and train it on samples with a varying number of observations between $1$ and $m$, $(n^i, \theta^i, x_1^i, \ldots, x^i_{n^i}) \sim \mathcal{U}(n;m) p(\theta) \prod_{j=1}^n p(x_j\vert \theta)$.

At inference time, given observations $x_1^o,\ldots,x_n^o$, the method approximately samples the target $p(\theta \vert x^o_1, \ldots, x^o_n)$ by partitioning the observations into subsets $X^o_1,\ldots,X^o_k$ and running annealed Langevin dynamics (\cref{alg:sampleuld}) with $c = \{X^o_1, \hdots, X^o_{k}\}$ and the approximate score
\begin{multline} \label{eq:partialscoreapprox}
    s_\psi(\theta, t, c) = \\ \frac{(1-n)(T-t)}{T} \nt \log p(\theta) + \sum_{j=1}^k s_\psi(\theta, t, X^o_j).
\end{multline}
The value of $m$ controls the trade-off between sample efficiency and error accumulation: for a given $m$, the method gets an approximate score by adding $k=\ceil{n/m}$ terms  as in \cref{eq:partialscoreapprox}, and requires on average $m/2$ simulator calls per training case. Therefore, low values of $m$ yield methods closer to \fnspe, while larger values yield methods closer to \nspe. In fact, we recover \fnspe for $m=1$  and \nspe for $m=n_{\max}$. This motivates finding the value of $m$ that achieves the best trade-off. We investigate this empirically in \cref{sec:exps}, where we observe that using low values of $m$ (greater than 1) often yields best results.

\section{Related Work}

Conditional score modeling has been used in many domains, including image super-resolution \cite{saharia2022image, li2022srdiff}, conditional image generation \cite{batzolis2021conditional}, and time series imputation \cite{tashiro2021csdi}. Concurrently with this work \citet{sharrock2022sequential} also studied its application for SBI, proposing a method akin to \nspe (they introduced the name \nspe, which we adopt here for consistency). Unlike our work that shows how \nspe can be extended to accommodate multiple observations, \citet{sharrock2022sequential} focus on the single-observation case and develop a sequential version of \nspe, which divides the learning process into stages and uses the learned posterior, instead of the prior, to draw parameters $\theta$ at each stage. This sequential approach of using the current posterior approximation to guide future simulations has its roots in the ABC literature \citep[e.g.][]{sisson2007sequential, blum2010nonlinear}. In the context of neural methods, it was originally proposed for NPE \cite{papamakarios2016fast}, and was later incorporated into NLE \cite{papamakarios2019sequential} and NRE \citep{hermans2020likelihoodfree}. Previous work has observed that sequential approaches often lead to performance improvements over their non-sequential counterparts \cite{greenberg2019automatic, lueckmann2021benchmarking} and are on par with active-learning methods for parameter selection \cite{durkan2018sequential}. While we do not explore sequential approaches, we note that \fnspe and \pfnspe are compatible with the sequential formulation of \citet{sharrock2022sequential}.

\citet{shi2022conditional} also studied the use of conditional diffusion models in the context of SBI, with the goal of reducing the number of diffusion steps required to obtain good results. To achieve this they extend the diffusion Schr\"{o}dinger bridge framework \cite{de2021diffusion} to perform conditional simulation. Their approach consists of a sequential training procedure where both the forward and backward processes are trained iteratively using score-matching techniques. They additionally propose to learn an observation-dependent reference $p_T(\theta \vert x)$ for the sampling process, replacing the standard Gaussian. Similarly to \citet{sharrock2022sequential}, they focus on the single observation case.

Finally, previous work by \citet{liu2022compositional} also proposed to compose diffusion models by adding their scores. However, they generate samples by directly plugging the composed score into the time-reverse diffusion of the noising process, which yields an incorrect sampling method (since the score of the true bridging densities does not follow this additive factorization, as explained in \cref{sec:diff_npe}). We address this issue by adopting a different sampler based on annealed Langevin dynamics. This solution was also concurrently proposed by \citet{du2023reduce}, who, among other things, proposed a method to compose diffusion models closely related to \fnspe.

\section{Empirical Evaluation} \label{sec:exps}

In this section, we empirically evaluate the proposed methods along with a number of baselines.
\Cref{sec:exps_multimodal,,sec:exps_systematic} show results comparing \fnspe, \pfnspe, \nspe, \npe, and \nre. We use 400 noise levels for methods based on score modeling, a normalizing flow with six Real NVP layers \cite{dinh2016density} for \npe, and the \nre method proposed by \citet{hermans2020likelihood} with HMC \cite{neal2011mcmc}. We provide the details for all the methods in \cref{app:details_arch}.

\cref{sec:opt_m,,sec:add_res_main} explore the robustness of \fnspe and \pfnspe for different design choices and hyperparameters, including the parameterization for the score network, parameters of the Langevin sampler, and the value of $m$ for \pfnspe. Unless specified otherwise, each method is given a budget of $10^4$ simulator calls, and optimization is carried out using Adam \cite{kingma2014adam} with the learning rate of $10^{-4}$ for a maximum of 20k epochs (using 20\% of the training data as a validation set for early stopping).

\begin{figure}[t]
  \centering
  \includegraphics[scale=0.3]{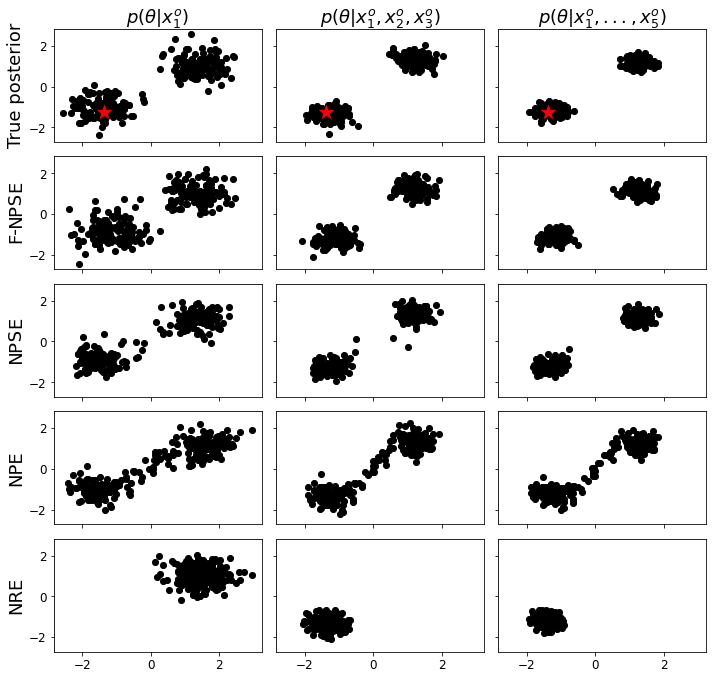}
  \caption{Samples from the approximate posterior obtained by each method. \npe and \nspe use $n_{\max}=5$. The true parameters $\theta$ used to generate $x_1^o, \hdots, x_5^o$ are shown in red in the first row.}
  \label{fig:mixture}
\end{figure}

\begin{figure*}[th!]
  \centering
  \includegraphics[scale=0.33]{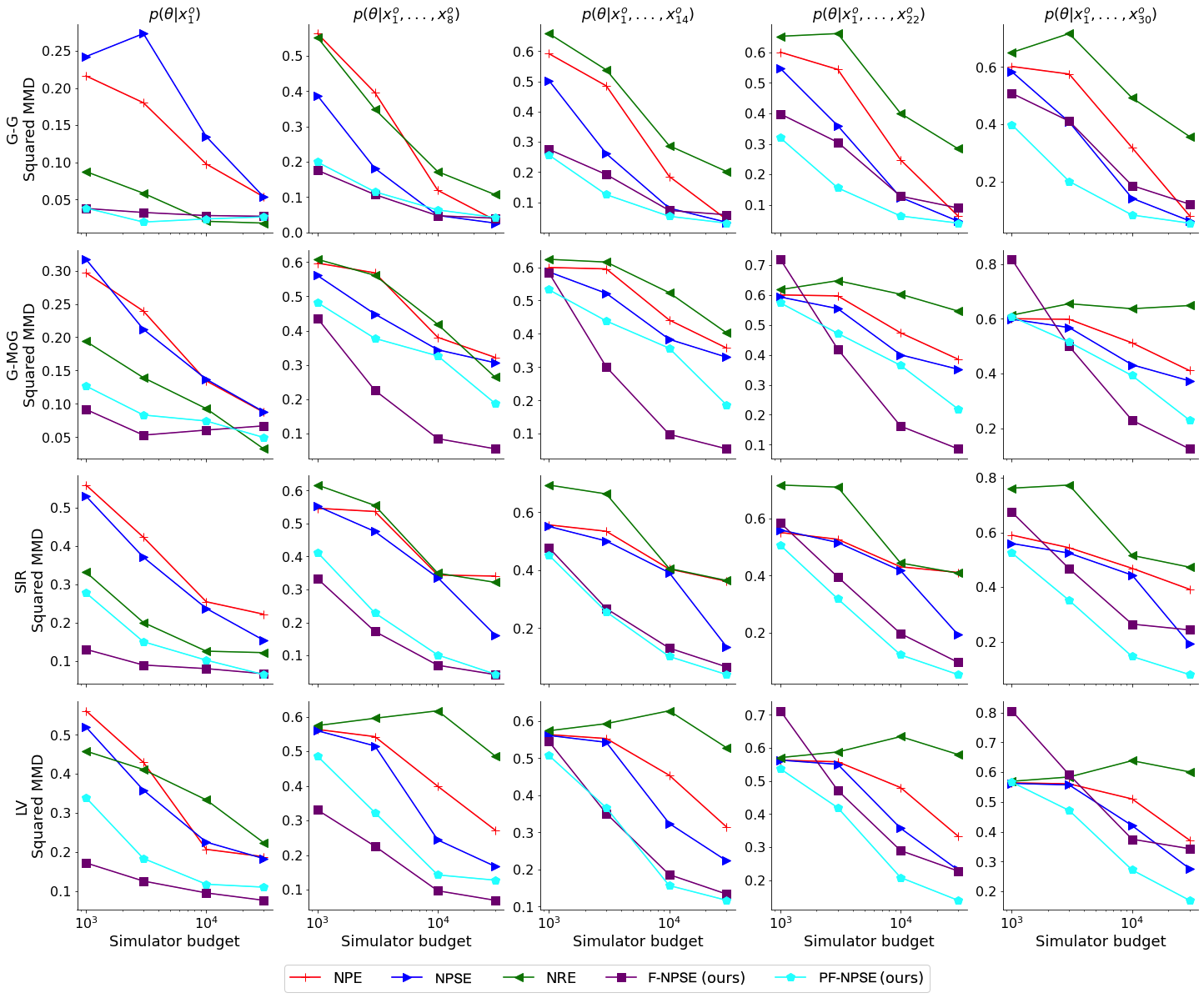}
  \caption{Squared MMD (lower is better) obtained by each method on different tasks. Plots show ``Squared MMD" ($y$-axis) vs. ``simulator budget used for training" ($x$-axis), and each line corresponds to a different method. Rows correspond to the different models considered (Gaussian/Gaussian, Gaussian/Mixture of Gaussians, SIR, Lotka--Volterra), and columns to the different number of conditioning observations available at inference time (1, 8, 14, 22, 30). We use $n_{\max}=30$ for \npe and \nspe and $m=6$ for \pfnspe.}
  \label{fig:res1}
\end{figure*}

\begin{figure*}[t]
  \centering
  \includegraphics[scale=0.37]{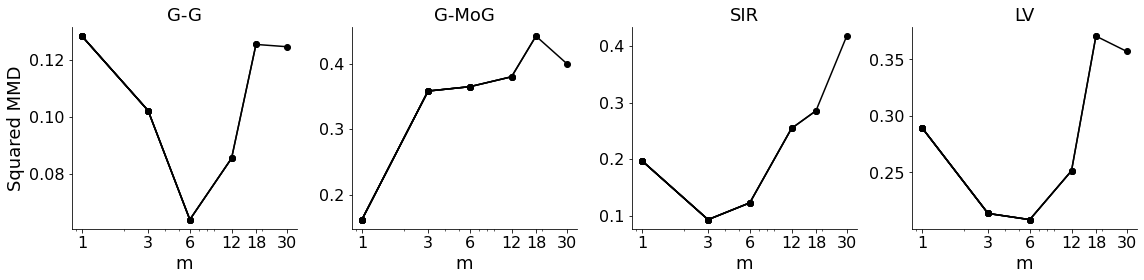}
  \caption{Squared MMD (lower is better) obtained by \pfnspe when estimating posterior distributions obtained by conditioning on 22 observations. Each plot corresponds to a different task, and shows ``Squared MMD" ($y$-axis) vs. ``$m$ (used for \pfnspe)" ($x$-axis).}
  \label{fig:res_m}
\end{figure*}

\subsection{Illustrative Multimodal Example} \label{sec:exps_multimodal}

We begin with a qualitative comparison on a 2-dimensional example with a multimodal posterior, where the prior and likelihood are given by $p(\theta) = \mathcal{N}(\theta \vert 0, I)$ and $p(x\vert \theta) = \frac{1}{2} \mathcal{N}(x\vert \theta, \frac{I}{2}) + \frac{1}{2} \mathcal{N}(x\vert -\theta, \frac{I}{2})$. We use $n_{\max}=5$ for \npe and \nspe. After training, we sample $\theta \sim p(\theta)$ and $x^o_1, \hdots, x^o_5 \overset{\mathrm{iid}}{\sim} p(x\vert \theta)$, and use each method to approximate the posterior distribution obtained by conditioning on subsets of $\{x^o_1, \hdots, x^o_5\}$ of different sizes. 

\Cref{fig:mixture} shows that \fnspe is able to capture both modes well for all subsets of observations considered, despite being trained using a single observation per training case. While NPE and \nspe also perform well, NRE fails to sample both modes, because HMC struggles to mix between modes, which is often an issue with MCMC samplers.

\subsection{Systematic Evaluation} \label{sec:exps_systematic}

We now present a systematic evaluation on four tasks typically used to evaluate SBI methods \cite{lueckmann2021benchmarking}, which are described in detail in \cref{app:models}.

\textbf{Gaussian/Gaussian (G-G).} A 10-dimensional model with a Gaussian prior $p(\theta) = \mathcal{N}(\theta\vert 0, I)$ and a Gaussian likelihood $p(x\vert \theta) = \mathcal{N}(x \vert \theta, \Sigma)$, with $\Sigma$ set to a diagonal matrix with elements increasing linearly from $0.6$ to $1.4$.

\textbf{Gaussian/Mixture of Gaussians (G-MoG).} A 10-dimensional model with a prior $p(\theta) = \mathcal{N}(\theta\vert 0, I)$ and a mixture of two Gaussians with a shared mean as the likelihood $p(x \vert \theta) = \frac{1}{2} \mathcal{N}(x\vert \theta, 2.25 \Sigma) + \frac{1}{2} \mathcal{N}(x\vert \theta, \frac{1}{9}\Sigma)$.

\textbf{Susceptible-Infected-Recovered (SIR).} A model used to track the evolution of a disease, modeling the number of individuals in each of three states: susceptible (S), infected (I), and recovered (R) \cite{harko2014exact}. The values for the variables S, I, R are governed by a non-linear dynamical system with two parameters, the contact rate and the recovery rate, which must be inferred from noisy observations of the number of individuals in each state at different times.

\textbf{Lotka--Volterra (LV).} A predator-prey model used in ecology to track the evolution of the number of individuals of two interacting species. The model consists of a non-linear dynamical system with four parameters, which must be inferred from noisy observations of the number of individuals of both species at different times.

We compare the methods' performance for different simulator call budgets $B\in\{10^3, 3\cdot 10^3, 10^4, 3\cdot 10^4\}$. We train using learning rates $\{10^{-3}, 10^{-4}\}$, keeping the one that yields the best validation loss for each method and simulator budget. We use $n_{\max}=30$ for \npe and \nspe and $m=6$ for \pfnspe. We train each method five times using different random seeds, and for each run we perform the evaluation using six different sets of parameters $\theta \sim p(\theta)$, not shared between runs.
After training, we generate observations by drawing $\theta \sim p(\theta)$ and $x_1^o, \hdots, x_{30}^o \overset{\mathrm{iid}}{\sim} p(x\vert \theta)$, and report each method's performance when approximating the posterior conditioned on subsets of $\{x_1^o, \hdots, x_{30}^o\}$ of different sizes. We report the squared Maximum Mean Discrepancy (MMD) \cite{gretton2012kernel} between the true posterior distribution and the approximations, using the Gaussian kernel with the scale determined by the median distance heuristic \cite{ramdas2015decreasing}. We also report results using classifier two-sample tests in \cref{app:c2st}.

We can draw several conclusions from \Cref{fig:res1}, which shows the average results. First, as expected, performance tends to improve for all methods as the simulator call budget is increased. Second, in most cases the top performing method is \fnspe or \pfnspe. In fact, \pfnspe typically outperforms all baselines, and is second to \fnspe when the number of conditioning observations is low (i.e., $n\leq 10$). The fact that \pfnspe with $m=6$ often outperforms both \fnspe and \nspe indicates that neither of the two extremes achieves the best trade-off between sample efficiency and accumulation of errors, and that methods in the spectrum interpolating between them are often preferred. We investigate this more thoroughly in \cref{sec:opt_m}. Finally, while \nre methods can naturally aggregate multiple observations at inference time, they also suffer from accumulation of errors. Results in \cref{fig:res1} suggest  that \fnspe and \pfnspe tend to scale better to large numbers of observations than \nre. We believe this may be due to  learning density ratios being a challenging problem, with new methods being actively developed \cite{rhodes2020telescoping, choi2022density}.

\subsection{Optimal Trade-off for \pfnspe} \label{sec:opt_m}

In this section, we study the performance of \pfnspe as a function of $m$. As explained in \cref{sec:ndpe}, the value of $m$ controls the trade-off between sample-efficiency and accumulation of errors. To investigate this trade-off, we fix $n_{\max} = 30$ and evaluate \pfnspe's performance for different values of $m\in\{1, 3, 6, 12, 18, 30\}$, which cover the spectrum between \fnspe and \nspe. \Cref{fig:res_m} shows results for approximating a posterior distribution  conditioned on $22$ observations, $x^o_1, \ldots, x^o_{22}$ (see \cref{app:results_m} for results for different numbers of conditioning observations). We see that the best performance is often achieved using small values of $m$, typically  $3$ or $6$, indicating that the methods located at the both ends of the spectrum, \fnspe and \nspe, are often suboptimal.

\subsection{Score Network and Langevin Sampler} \label{sec:add_res_main}

We now investigate the robustness of the proposed methods to different design/hyperparameter choices. Specifically, we study the effect of constraining the score network to output a conservative vector field, by taking the score to be the gradient of a scalar-valued network \cite{salimans2021should}, and of the choice of the step-size and the number of steps per noise level in the Langevin sampler. We show the results in \cref{app:cons,,app:ldsampler}. Overall we find that our methods are robust to different choices: \cref{fig:res_cons} shows that using a conservative parameterization for the score network does not have a considerable effect on performance, and \cref{fig:ldsampler} shows that \pfnspe is robust to changes in the Langevin sampler, as long as it performs a sufficient number of steps (typically 5--10) to converge for each noise level.

\subsection{Demonstration: Weinberg Simulator}

\begin{figure}[th]
  \centering
    \includegraphics[scale=0.275]{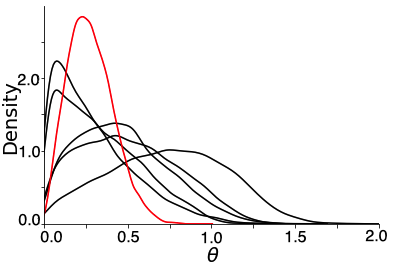}
    \includegraphics[scale=0.275]{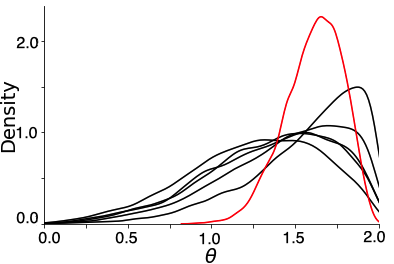}
  \caption{Approximations obtained by \fnspe on the Weinberg task. Left: $\theta^*=0.3$. Right: $\theta^*=1.7$. Black lines show approximations obtained for $p(\theta \vert x_i)$ for $i=1,\ldots,5$, and red lines shows the approximation for $p(\theta \vert x_1, \ldots, x_5)$. We note that the approximations obtained are constrained to the $[0, 2]$ interval (since we reparameterize the model, see \cref{app:models}). The fact that the plots extend beyond these limits is an artifact due to the use of a kernel density estimator.}
  \label{fig:wein}
\end{figure}

We conclude our evaluation with a demonstration of \fnspe with the Weinberg simulator, introduced as a benchmark by \citet{louppe2017adversarial}. The simulator models high-energy particle collisions, and the goal is to estimate the Fermi constant given observations for the scattering angle. We use the simplified simulator from \citet{activesciencing} with a uniform prior $p(\theta) = \mathcal{U}(0, 2)$. To evaluate, we fix the parameters $\theta^*$ (we consider $\theta^*=0.3$ and $\theta^*=1.7$), draw $x_1, \ldots, x_5 \overset{\mathrm{iid}}{\sim} p(x\vert \theta^*)$, and estimate the posteriors $p(\theta \vert x_i)$ for $i=1,\ldots,5$ and $p(\theta \vert x_1, \ldots, x_5)$. \Cref{fig:wein} shows the results. Posteriors conditioned on individual observations and on all five observations are shown in black and red, respectively. As expected, as the number of observations is increased the posterior returned by \fnspe concentrates around the true parameter value $\theta^*$.

\section{Conclusion and Limitations}

We studied the use of conditional score modeling for SBI, proposing several methods that can aggregate information from multiple observations at inference time while requiring a small number of simulator calls to generate each training case. We achieve this using a novel construction for the bridging densities, which allows us to simply aggregate the scores to approximately sample distributions conditioned on multiple observations. We presented an extensive empirical evaluation, which shows promising results for our methods when compared against other approaches able to aggregate an arbitrary number of observations at inference time.

As presented, \fnspe and \pfnspe use annealed Langevin dynamics to generate samples. This requires choosing step-sizes $\delta_t$, the number of steps $L$ per noise level, and has complexity $\mathcal{O}(LT)$. \Cref{app:samplingnould} presents an alternative sampling method that can be used with \fnspe and \pfnspe which does not use Langevin dynamics, does not have any hyperparameters, and runs in $\mathcal{O}(T)$ steps.
While our evaluation of this sampling method shows promising results (\cref{app:ldsampler}), we observe that sometimes it underperforms annealed Langevin dynamics. We believe that further study of alternative sampling algorithms is a promising direction for improving \fnspe and \pfnspe.

\section*{Acknowledgements}

The authors would like to thank Francisco Ruiz and Bobby He for helpful comments and suggestions.

\bibliography{references}
\bibliographystyle{icml2023}

\newpage
\appendix
\onecolumn

\section{Models Used} \label{app:models}

\paragraph{Multimodal posterior from \cref{sec:exps_multimodal}.}

We consider $\theta \in \mathbb{R}^{2}$ and $x \in \mathbb{R}^{2}$, and the prior and likelihood are given by
\begin{equation}
    p(\theta) = \mathcal{N}(\theta \vert 0, I) \quad \mbox{and} \quad p(x\vert \theta) = \frac{1}{2}\mathcal{N}\left(x\big\vert \theta, \frac{I}{2}\right) + \frac{1}{2} \mathcal{N}\left(x\big\vert -\theta, \frac{I}{2}\right).
\end{equation}

\paragraph{Gaussian/Gaussian (G-G).} This model was adapted from \citet{lueckmann2021benchmarking}. We consider $\theta \in \mathbb{R}^{10}$ and $x \in \mathbb{R}^{10}$, and the prior and likelihood given by
\begin{equation}
p(\theta) = \mathcal{N}(\theta\vert 0, I)\quad \mbox{and} \quad p(x\vert \theta) = \mathcal{N}(x \vert \theta, \Sigma),
\end{equation}
where $\Sigma$ is a diagonal matrix with elements increasing linearly from $0.6$ to $1.4$.

\paragraph{Gaussian/Mixture of Gaussians (G-MoG).} A model similar to this one was originally proposed by \citet{sisson2007sequential}, and was later widely adopted in the SBI literature \cite{lueckmann2021benchmarking}. We consider $\theta \in \mathbb{R}^{10}$ and $x \in \mathbb{R}^{10}$, and the prior and likelihood given by
\begin{equation}
    p(\theta) = \mathcal{N}(\theta\vert 0, I) \quad \mbox{and} \quad p(x \vert \theta) = \frac{1}{2}\mathcal{N}\left(x\big\vert \theta, 2.25 \Sigma \vphantom{\frac{1}{9}}\right) + \frac{1}{2} \mathcal{N}\left(x\big\vert \theta, \frac{1}{9}\Sigma\right),
\end{equation}
where $\Sigma$ is a diagonal matrix with elements increasing linearly from $0.6$ to $1.4$.

\paragraph{Susceptible-Infected-Recovered (SIR).} This model describes the transmission of a disease through a population of size $N$, where each individual can be in one of three states: susceptible $S$, infectious $I$, or recovered $R$. We follow the description from \citet[\S T9]{lueckmann2021benchmarking}. The model has two parameters, the contact rate $\beta$ and the transmission rate $\gamma$, i.e. $\theta = (\beta, \gamma)$. The simulator numerically simulates the dynamical system given by
\begin{equation}
\begin{split}
    \frac{\mathrm{d} S}{\mathrm{d} t} & = -\beta \frac{S I}{N}\\
    \frac{\mathrm{d} I}{\mathrm{d} t} & = \beta \frac{S I}{N} - \gamma I\\
    \frac{\mathrm{d} R}{\mathrm{d} t} & = \gamma I\\
\end{split}
\end{equation}
for $160$ seconds, starting from the initial conditions where a single individual is infected and the rest susceptible. The prior over the parameters is given by
\begin{equation}
    p(\beta) = \mathrm{LogNormal}(\log(0.4), 0.5) \quad \mbox{and} \quad p(\gamma) = \mathrm{LogNormal}(\log(0.125), 0.2).
\end{equation}
The observations are 10-dimensional vectors corresponding to noisy observations of the number of infected people at 10 evenly-spaced times, $p(x_i) = \mathcal{B}\left(1000, \frac{I_i}{N}\right)$, where $\mathcal{B}$ denotes the binomial distribution and $I_i$ the number of infected subjects at the corresponding time.

\paragraph{Lotka--Volterra (LV).} This model describes the evolution of the number of individual of two interacting species, typically a prey and a predator. We follow the description from \citet[\S T10]{lueckmann2021benchmarking}. The model has four parameters $\alpha, \beta, \gamma, \delta$. Using $X$ and $Y$ to represent the number of individuals in each species, the simulator numerically simulates the dynamical system given by
\begin{equation}
\begin{split}
    \frac{\mathrm{d} X}{\mathrm{d} t} & = \alpha X - \beta XY\\
    \frac{\mathrm{d} Y}{\mathrm{d} t} & = -\gamma Y + \delta XY.\\
\end{split}
\end{equation}
The system is simulated for $20$ seconds, starting from $X_0=30$ and $Y_0=1$. The prior over the parameters is given by
\begin{equation}
\begin{split}
    p(\alpha) & = \mathrm{LogNormal}(-0.125, 0.5)\\
    p(\beta) & = \mathrm{LogNormal}(-3, 0.5)\\
    p(\gamma) & = \mathrm{LogNormal}(-0.125, 0.5)\\
    p(\delta) & = \mathrm{LogNormal}(-3, 0.5).
\end{split}
\end{equation}
The observations consist of 20-dimensional vectors corresponding to noisy observations of the number of members of each species at 10 evenly-spaced times, $p(x_{i, X}) = \mathrm{LogNormal}(\log(X_i), 0.1)$ and $p(x_{i, Y}) = \mathrm{LogNormal}(\log(Y_i), 0.1)$.

\paragraph{Weinberg simulator.} This task was originally proposed by \citet{louppe2017adversarial}. We use the simplified simulator of \citet{activesciencing} with a uniform prior $p(\theta) = \mathcal{U}(0, 2)$.

\paragraph{Reparameterizing models.} For all tasks, we reparameterize the models to have a standard Normal prior. This is simple to do with the priors commonly used by these models.

\section{Details for each Method} \label{app:details_arch}

All methods based on denoising diffusion models/score modeling use $T=400$.

\subsection{\fnspe}

Our implementation of the score network $s_\psi(\theta, t, x)$ used by \fnspe has three components:
\begin{itemize}
    \item An MLP with 3 hidden layers that takes $\theta$ as input and outputs an embedding $\theta_\mathrm{emb}$,
    \item An MLP with 3 hidden layers that takes $x$ as input and outputs an embedding $x_\mathrm{emb}$,
    \item An MLP with 3 hidden layers that takes $[\theta_\mathrm{emb}, x_\mathrm{emb}, t_\mathrm{emb}]$ as input, where $t_\mathrm{emb}$ is a positional embedding obtained as described by \citet{vaswani2017attention}, and outputs an estimate of the score. We parameterize the score in terms of the noise variables \cite{ho2020denoising, luo2022understanding}.
\end{itemize}
All MLPs use residual connections \cite{he2016deep} and normalization layers \citep[LayerNorm,][]{ba2016layer} throughout.

Running \cref{alg:sampleuld} to generate samples using the trained score network requires choosing step sizes $\delta_t$ and the number of Langevin steps $L$ for each noise level $\gamma_t$. We use $L=5$ and $\delta_t = 0.3 \frac{1-\alpha_t}{\sqrt{\alpha_t}}$, where $\alpha_1 = \gamma_1$ and $\alpha_t = \frac{\gamma_t}{\gamma_{t-1}}$ for $t= 2, \hdots, T-1$.

\subsection{\pfnspe}

The architecture for the score network $s_\psi(\theta, t, X)$, where $X$ is a set of observations with a number of elements between $1$ and $m$, is similar to the one used for \fnspe, with two differences. First, each individual observation $x_i\in X$ produces and embedding $x_{i,\mathrm{emb}}$, and the final embedding $X_\mathrm{emb}$ is obtained by averaging the individual embeddings. Second, the final MLP takes as inputs $[\theta_\mathrm{emb}, X_\mathrm{emb}, t_\mathrm{emb}, n_\mathrm{emb}]$, where $n_\mathrm{emb} \in \{0, 1\}^m$ is a 1-hot encoding of the number of observations in the set $X$ (between $1$ and $m$). The Langevin sampler uses the parameters described above.

\subsection{\nspe}

The architecture is the same as the one for \pfnspe, with $m = n_{\max}$. Since the method corresponds to a direct application of conditional diffusion models, we use the standard sampler for diffusion models \cite{ho2020denoising}, which consists of applying $T$ Gaussian transitions, with no tunable parameters.

\subsection{\npe}

We use an implementation of NPE methods based on flows able to handle sets of observations of any size $n\in\{1, 2, \hdots n_{\max}\}$. The flow can be expressed as $q_\psi(\theta \vert x_1, \hdots, x_n, n)$. Following \citet{chan2018likelihood} and \citet[\S2.4]{radev2020bayesflow}, we use an exchangeable neural network to process the observations $x_1, \hdots, x_n$. Specifically, we use an MLP with 3 hidden layers to generate an embedding $x_{i,\mathrm{emb}}$ for each observation $x_i$. We then compute the mean embedding across observations $x_\mathrm{emb} = \frac{1}{n} \sum_i^n x_{i,\mathrm{emb}}$, which we use as input for the conditional flow. Finally, we model the flow $q_\psi(\theta \vert x_1, \hdots, x_n, n) = q_\psi(\theta \vert x_\mathrm{emb}, n_\mathrm{emb})$, where $n_\mathrm{emb} \in \{0, 1\}^{n_{\max}}$ is a 1-hot encoding of the number of observations (between $1$ and $n_{\max}$).

We use 6 Real NVP layers \cite{dinh2016density} for the flow, each one consisting of MLPs with three hidden layers. As for the other methods, we use residual connections throughout.

\subsection{\nre}

We use the \nre method proposed by \citet{hermans2020likelihood}. Simply put, the network receives a pair $(\theta, x)$ as input, and is trained to classify whether $(\theta, x) \sim p(\theta, x)$ or $(\theta, x) \sim p(\theta) p(x)$. The optimal classifier learnt this way can be used to compute the ratio $\frac{p(x\vert \theta)}{p(x)}$. Our classifier consists of three components: two linear layers to compute embeddings for $x$ and $\theta$, and a three-layer MLP (using residual connections) that takes the concatenated embeddings as input. We tried using a larger network, but got slightly worse results.

To sample the target distribution using the learned ratio we use HMC \cite{neal2011mcmc}. More specifically, we use NumPyro's \cite{phan2019composable} implementation of the No-U-Turn-Sampler \cite{hoffman2014no}, a variant of HMC that automatically sets the number of leapfrog steps.

\section{Derivation of Posterior Factorization} \label{app:derfact}

The factorization for the posterior distribution $p(\theta \vert x_{1}, \hdots, x_{n})$ is obtained applying Bayes rule twice:
\begin{align}
    p(\theta \vert x_{1}, \hdots, x_{n}) & \propto p(\theta) p(x_1, \hdots, x_n \vert \theta) & \mbox{(Bayes rule)}\\
    & = p(\theta) \prod_{j=1}^n p(x_{j} \vert \theta)\\
    & \propto p(\theta) \prod_{j=1}^n \frac{p(\theta \vert x_{j})}{p(\theta)} & \mbox{(Bayes rule)}\\
    & = p(\theta)^{1-n} \prod_{j=1}^n p(\theta \vert x_{j}).
\end{align}

\section{Alternative Sampling Method Without Unadjusted Langevin Dynamics} \label{app:samplingnould}

The sampling process described in \cref{alg:sampleuld} requires choosing step-sizes $\delta_t$ and the number of steps $L$ per noise level, and has complexity $\mathcal{O}(LT)$. This section introduces a different method to approximately sample the target $p(\theta \vert x_1, \hdots, x_n)$, which does not use Langevin dynamics, does not require step-sizes $\delta_t$, and runs in $\mathcal{O}(T)$ steps. The approach is based on the formulation by \citet{sohl2015deep} to use diffusion models to approximately sample from a product of distributions. The final method, shown in \cref{alg:samplenould}, involves sampling $T$ Gaussian transitions with means and variances computed using the trained score network, and reduces to the typical approach to sampling diffusion models \cite{ho2020denoising} for $n=1$. Each iteration of the algorithm consists of four main steps: Lines 6 and 7 compute the transition's mean from the learned scores of $p_t(\theta \vert x_i)$, line 8 computes the variance of the transition, line 9 corrects the mean to account for the prior term, and line 10 samples from the resulting Gaussian transition.

\begin{algorithm}[ht]
\caption{Sampling without unadjusted Langevin}
\label{alg:samplenould}
\begin{algorithmic}[1]
\STATE {\bfseries Input:} Score network $s_\psi(\theta, t, x)$, reference $p_T(\theta)$, observations $\{x_1, \hdots, x_n\}$, noise levels $\gamma_1, \hdots, \gamma_T$
\STATE $\alpha_1 \coloneqq\gamma_1$ and $\alpha_t \coloneqq \frac{\gamma_t}{\gamma_{t-1}}$ for $t=2,\hdots,T-1$
\STATE $\theta \sim p_T(\theta)$ \hfill $\triangleright$ Sample reference
\FOR{$t = T-1, T-2, \hdots, 1$}
    \STATE $\mu_t = \frac{1}{n - \alpha_t(n-1)} \left[\sum_j \left( \frac{\theta}{\sqrt{\alpha_t}} + \frac{(1 - \alpha_t)}{\sqrt{\alpha_t}} s_\psi(\theta, t, x_j) \right) - (n-1) \sqrt{\alpha_t} \theta \right]$ \hfill $\triangleright$ Compute transition's mean from scores \\ \hfill
    \STATE $\mu_t \mathrel{+}= \frac{\sigma_t^2(1-n)(T-t)}{T} \nabla_\theta \log p(\theta)$ \hfill $\triangleright$ Prior correction term \\ \hfill
    \STATE $\sigma^2_t = \frac{1-\alpha_t}{n - \alpha_t(n-1)}$ \hfill $\triangleright$ Compute transition's variance \\ \hfill
    \STATE $\theta \sim \mathcal{N}(\theta\vert \mu_t, \sigma^2_t I)$ \hfill $\triangleright$ Sample transition
\ENDFOR
\end{algorithmic}
\end{algorithm}

We now give the derivation for the sampling method shown in \cref{alg:samplenould}. In short, the derivation uses the formulation of score-based methods as diffusions, and has 3 main steps: (1) using the scores of $p_t(\theta\vert x)$ to compute the Gaussian transition kernels of the corresponding diffusion process for the target $p(\theta \vert x)$; (2) composing $n$ Gaussian transitions corresponding to the $n$ diffusions of $p_t(\theta\vert x_1), \hdots, p_t(\theta\vert x_n)$  \citep[this is based on][]{sohl2015deep}; and (3) adding a correction for the prior term  \citep[also based on][]{sohl2015deep}. We note that steps 2 and 3 require some approximations. We believe a thorough analysis of these approximations would be useful in understanding when the sampling method can be expected to work well. For clarity, we use [A] to indicate when the approximations are introduced/used.

\paragraph{(1) Connection between score-based methods and diffusion models.} We begin by noting that score-based methods can be equivalently formulated as diffusion models, where the mean of Gaussian transitions that act as denoising steps are learned instead of the scores. Specifically, letting $\alpha_1=\gamma_1$ and $\alpha_t = \frac{\gamma_t}{\gamma_{t-1}}$ for $t=2,\hdots,T-1$, the learned model is given by a sequence of Gaussian transitions $p_t(\theta_{t-1} \vert \theta_{t}, x) = \mathcal{N}(\theta_{t-1} \vert \mu_{\psi}(\theta_{t}, t, x), 1-\alpha_t)$ trained to invert a sequence of noising steps given by $q_t(\theta_{t}\vert \theta_{t-1}) = \mathcal{N}(\theta_{t} \vert \sqrt{\alpha_t} \, \theta_{t-1}, (1-\alpha_{t}) I)$. The connection between diffusion models and score-based methods comes from the fact that the optimal means and scores are linearly related by \cite{luo2022understanding}
\begin{equation} \label{eq:linrel}
    \mu_\psi(\theta, t, x) = \frac{1}{\sqrt{\alpha_t}} \theta + \frac{1 - \alpha_t}{\sqrt{\alpha_t}} s_\psi(\theta, t, x).
\end{equation}

\paragraph{(2) Approximately composing $n$ diffusions.} To simplify notation, we use a superscript $j$ to indicate distributions that are conditioned on $x_j$ (e.g.\ $p^j_{t}(\theta_{t}) = p_{t}(\theta_{t} \vert x_j)$). Assume we have transition kernels $p_t^j(\theta_{t-1} \vert \theta_{t})$ that exactly reverse the forward kernels $q(\theta_{t}\vert \theta_{t-1})$ [Assumption A1], meaning that $p^j_{t-1}(\theta_{t-1}) = \int d\theta_t \, p^j_{t}(\theta_t) \, p_t^j(\theta_{t-1} \vert \theta_t)$, or equivalently $p^j_{t}(\theta_t) \, p_t^j(\theta_{t-1} \vert \theta_t) = p^j_{t-1}(\theta_{t-1}) \, q_t(\theta_{t} \vert \theta_{t-1})$. Our goal is to find a transition kernel $\tilde p_t(\theta_{t-1} \vert \theta_t)$ that satisfies
\begin{equation} \label{eq:tildep}
\tilde p_{t-1}(\theta_{t-1}) = \int d\theta_t \, \tilde p_{t}(\theta_t) \, \tilde p_t(\theta_{t-1} \vert \theta_t),
\end{equation}
where $\tilde p_{t}(\theta_t) = \frac{1}{Z_t} \prod_{j}^n p^j_t(\theta_t)$. This is closely related to our formulation in \cref{sec:ndpe}, since our definition for the bridging densities involves the product $\prod_j^n p_t(\theta \vert x_j)$. The condition from \cref{eq:tildep} can be re-written as
\begin{align}
p^1_{t-1}(\theta_{t-1}) & = \int d\theta_t \, p^1_{t}(\theta_{t}) \frac{p^2_{t}(\theta_{t})}{p^2_{t-1}(\theta_{t-1})} \cdots \frac{p^n_{t}(\theta_{t})}{p^n_{t-1}(\theta_{t-1})} \frac{Z_{t-1}}{Z_t} \tilde p_t(\theta_{t-1} \vert \theta_t)\\
& = \int d\theta_t \, p^1_{t}(\theta_{t})
\textcolor{blue}{
\frac{q_t(\theta_{t} \vert \theta_{t-1})}{p_t^2(\theta_{t-1} \vert \theta_t)} \cdots
\frac{q_t(\theta_{t} \vert \theta_{t-1})}{p_t^n(\theta_{t-1} \vert \theta_t)}
\frac{Z_{t-1}}{Z_t} \tilde p_t(\theta_{t-1} \vert \theta_t)} & \mbox{[A1]}. \label{eq:decsd}
\end{align}
One way to satisfy \cref{eq:decsd} is by setting $\tilde p_t(\theta_{t-1} \vert \theta_t)$ so that the term in blue above is equal to $p_t^1(\theta_{t-1} \vert \theta_t)$ \cite{sohl2015deep}. That is,
\begin{equation}\label{eq:unnormtransition}
\tilde p_t(\theta_{t-1} \vert \theta_t) = p_t^1(\theta_{t-1} \vert \theta_t) \frac{Z_{t}}{Z_{t-1}}
\frac{p_t^2(\theta_{t-1} \vert \theta_t)}{q_t(\theta_{t} \vert \theta_{t-1})} \cdots
\frac{p_t^n(\theta_{t-1} \vert \theta_t)}{q_t(\theta_{t} \vert \theta_{t-1})}.
\end{equation}
However, the resulting $\tilde p_t(\theta_{t-1} \vert \theta_t)$ may not be a normalized distribution \cite{sohl2015deep}. Following \citet{sohl2015deep}, we propose to use the corresponding normalized distribution defined as $\tilde p_t^N(\theta_{t-1} \vert \theta_t) \propto \tilde p_t(\theta_{t-1} \vert \theta_t)$ [Assumption A2]. Given that \cref{eq:unnormtransition} corresponds to the product of Gaussian densities, the resulting normalized transition is also Gaussian, with mean and variance given by 
\begin{equation} \label{eq:meanvarder}
\mu_t = \frac{\sum_j \mu_{jt} - (n-1) \sqrt{\alpha_t} \theta}{n - \alpha_t(n-1)}  
\quad \mbox{and} \quad
\sigma^2_t = \frac{1-\alpha_t}{n - \alpha_t(n-1)}.
\end{equation}

\paragraph{(3) Prior correction term.} The formulation above ignores the fact that the bridging densities defined in \cref{eq:approachbr} involve the prior $p(\theta)$. We use the method proposed by \citet{sohl2015deep} to correct for this, which involves adding the term $\frac{\sigma_t^2(1-n)(T-t)}{T} \nabla_\theta \log p(\theta)$ to the mean $\mu_t$ from \cref{eq:meanvarder}. The derivation for this is similar to the one above, and also requires setting the resulting transition kernel to the normalized version of an unnormalized distribution \cite{sohl2015deep}.

As mentioned previously, this derivation uses two assumptions/approximations. [A1] assumes that the learned score function/reverse diffusion approximately reverses the noising process, which is reasonable if the forward kernels $q_t$ add small amounts of noise per step (equivalently, if the noise levels $\gamma_1, \hdots, \gamma_T$ increase slowly). [A2] assumes that the normalized version of $\tilde p_t(\theta_{t-1} \vert \theta_t)$, given by $\tilde p^N_t(\theta_{t-1} \vert \theta_t)$, approximately satisfies \cref{eq:decsd}.

\newpage

\section{Additional Results} \label{app:addres}

\subsection{Classifier Two-sample Test} \label{app:c2st}

\begin{figure*}[th!]
  \centering
  \includegraphics[scale=0.33]{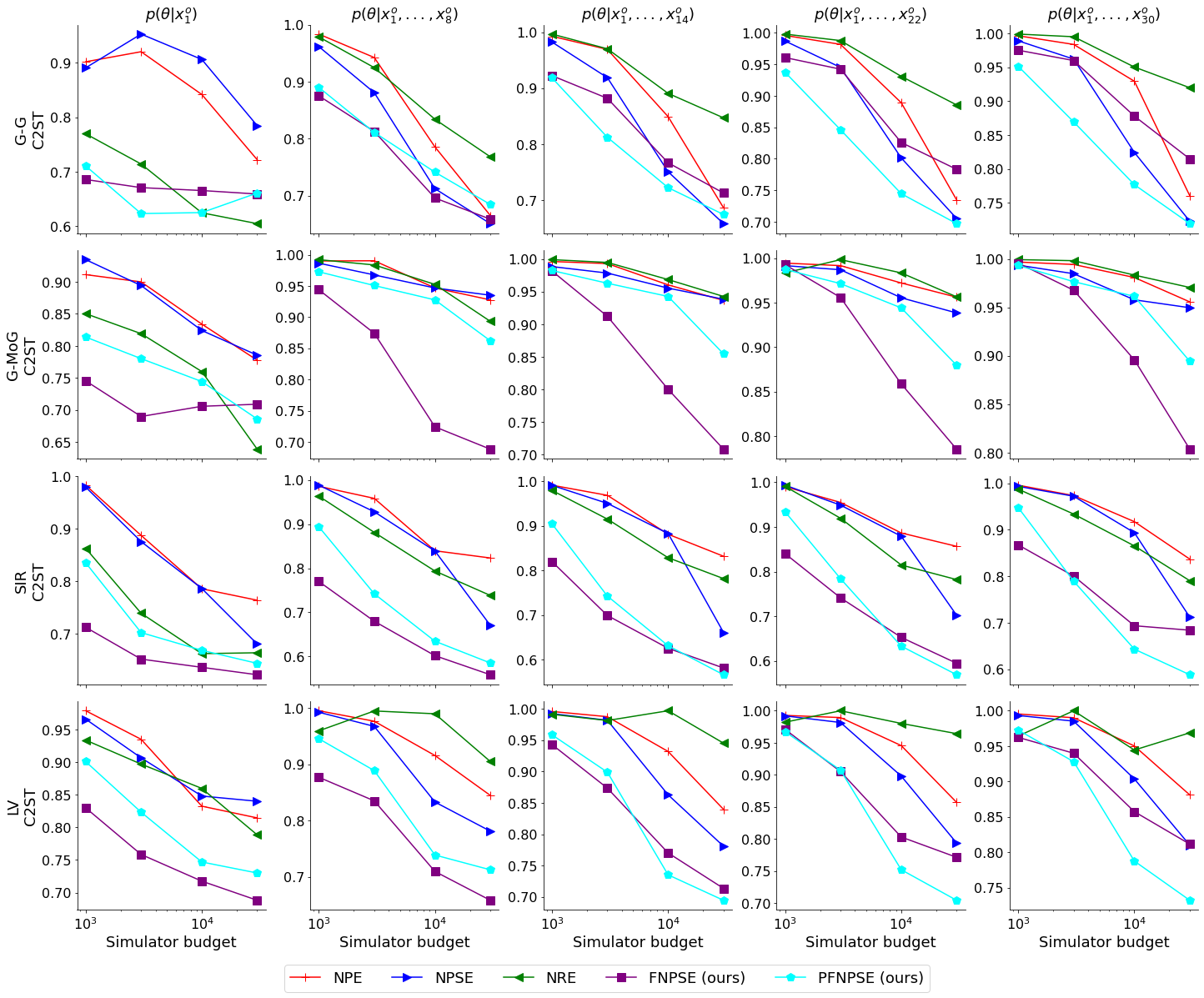}
  \caption{Classifier two-sample test scores (C2ST, lower is better) obtained by each method on different tasks. Plots show ``C2ST" ($y$-axis) vs. ``simulator budget used for training" ($x$-axis), and each line corresponds to a different method. Rows correspond to the different models considered (Gaussian/Gaussian, Gaussian/Mixture of Gaussians, SIR, Lotka--Volterra), and columns to the different number of conditioning observations available at inference time (1, 8, 14, 22, 30). We use $n_{\max}=30$ for \npe and \nspe and $m=6$ for \pfnspe.}
  \label{fig:c2st}
\end{figure*}

\Cref{fig:c2st} shows the evaluation of  different methods using the classifier two-sample test (C2ST)  \cite{friedman2003multivariate, lueckmann2021benchmarking}. Essentially, we train a classifier (a feed-forward network with three layers) to discriminate between samples coming from the trained model and the true posterior. The metric reported is the accuracy of the trained classifier on the test set. The metric ranges between $0.5$ and $1$, with the former indicating a very good posterior approximation (the classifier cannot discriminate between true samples and the ones provided by the model) and the latter a poor one (the classifier can perfectly discriminate between true samples and the ones provided by the model). These C2ST scores were computed from the same runs as the results in \cref{fig:res1}, so all the training details are the same.

\clearpage
\newpage

\subsection{Performance of \pfnspe for Different $m$} \label{app:results_m}

\begin{figure*}[th!]
  \centering
  \includegraphics[scale=0.3]{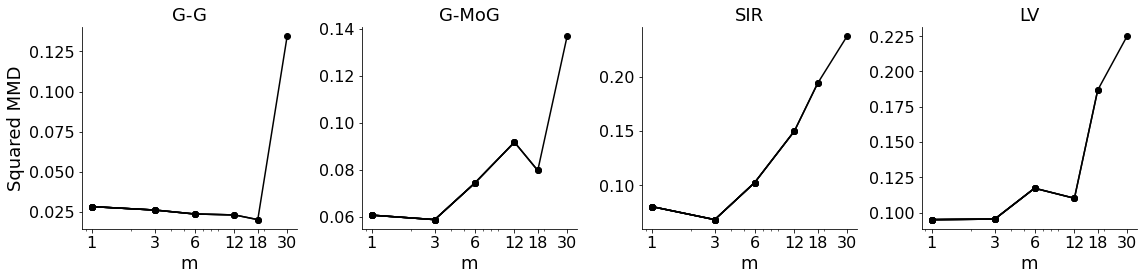}
  
  \includegraphics[scale=0.3]{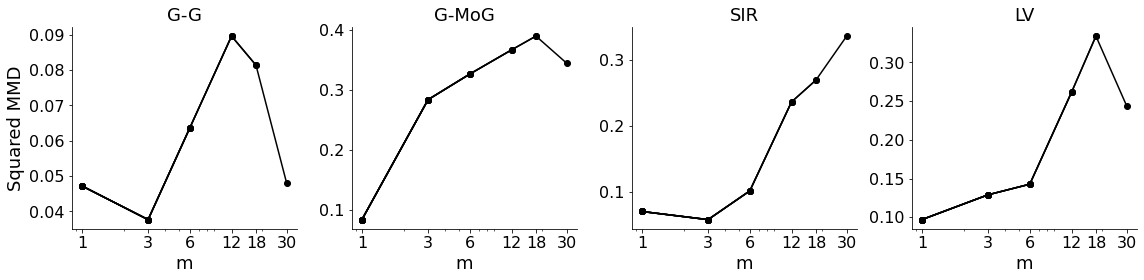}
  
  \includegraphics[scale=0.3]{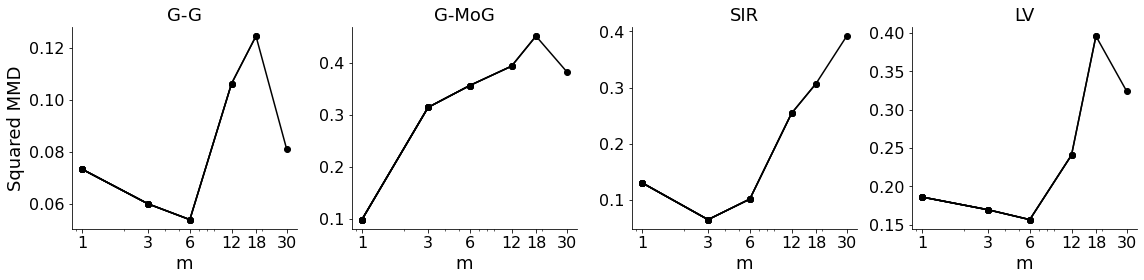}
  
  \includegraphics[scale=0.3]{m_study/m_10k_22.png}
  
  \includegraphics[scale=0.3]{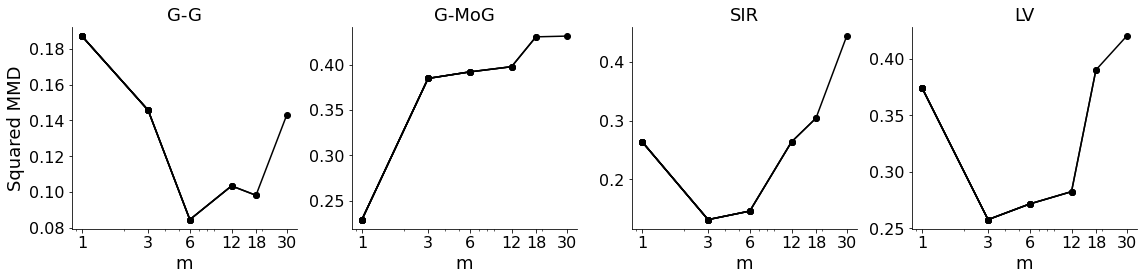}
  \caption{Plots show squared MMD (lower is better). Each column corresponds to a task (G-G, G-MoG, SIR, LV), and each row corresponds to approximating distributions obtained by conditioning on a different number of observations. From top to bottom, the number of conditioning observations we use is $1$, $8$, $14$, $22$, and $30$.}
  \label{fig:res_m_app}
\end{figure*}

\Cref{fig:res_m_app} shows the results obtained using \pfnspe for $m \in \{1, 3, 6, 12, 18, 30\}$ to estimate the posterior distributions obtained by conditioning on a different number of observations in $\{1, 8, 14, 22, 30\}$. As pointed out in the main text, values of $m$ between $3$ and $6$ often lead to the best results.

\clearpage
\newpage

\subsection{Conservative and Non-conservative Parameterization for the Score Network} \label{app:cons}

\begin{figure*}[th!]
  \centering
  \includegraphics[scale=0.33]{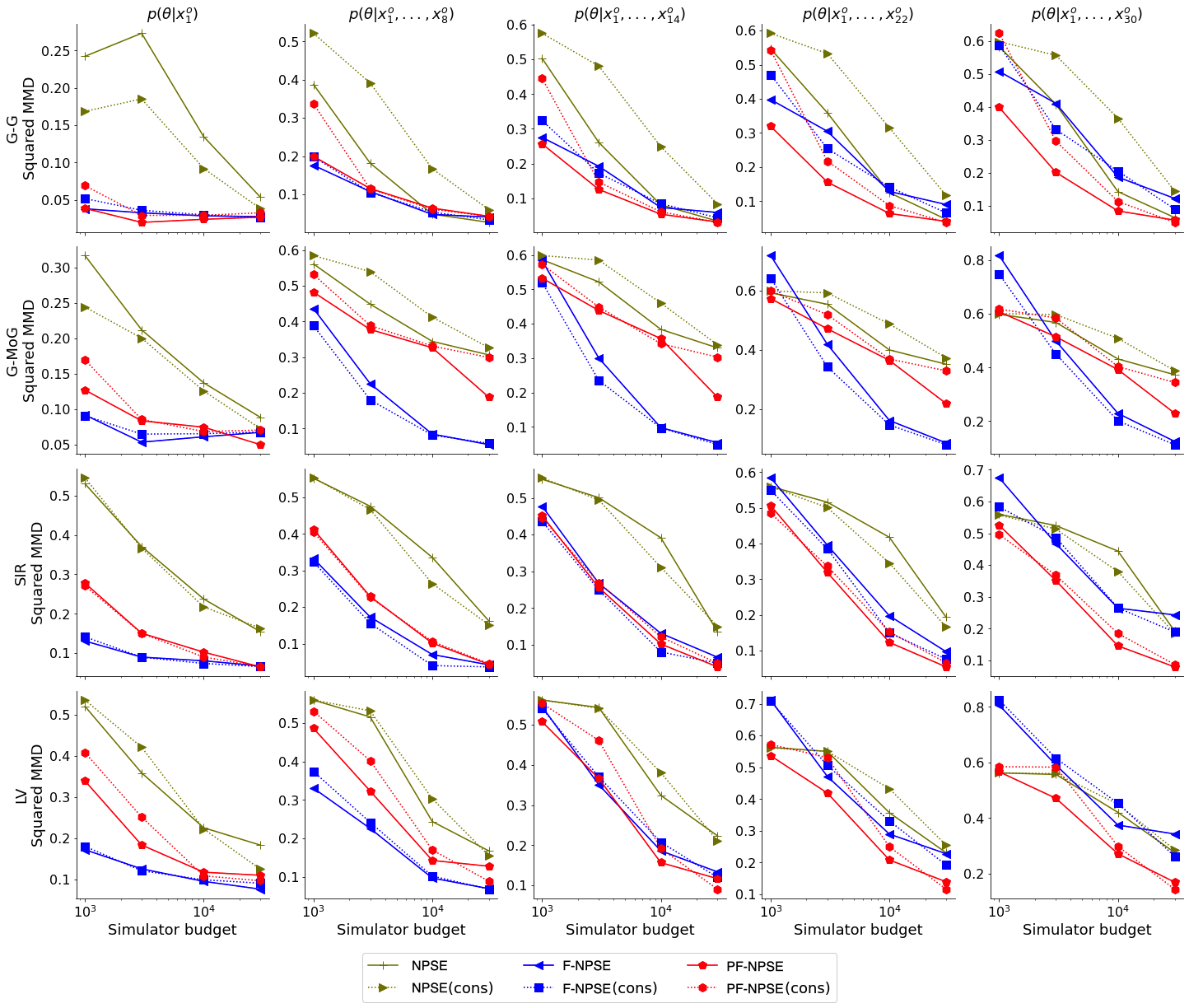}
  
  \caption{Squared MMD (lower is better) achieved by different methods on different tasks. Each row corresponds to a different model considered (Gaussian-Gaussian, Gaussian-Mixture of Gaussians, SIR, Lotka--Volterra), and each column to a different number of conditioning observations available at inference time (1, 8, 14, 22, 30). The \textit{(cons)} identifier corresponds to methods using the conservative parameterization for the score network.}
  \label{fig:res_cons}
\end{figure*}

\Cref{fig:res_cons} compares the results obtained by score-modeling methods (\nspe, \fnspe, and \pfnspe) using different parameterizations for the score network. We consider the unconstrained parameterization typically used, where the score network outputs a vector of the same dimension as $\theta$, as well as a conservative parameterization, where the network outputs a scalar, and the score is obtained by computing its gradient. The figure shows that the choice of parameterization does not tend to have a substantial effect on performance.

\clearpage
\newpage

\subsection{Langevin Sampler Parameters} \label{app:ldsampler}

\begin{figure*}[th!]
  \centering
  \includegraphics[scale=0.33]{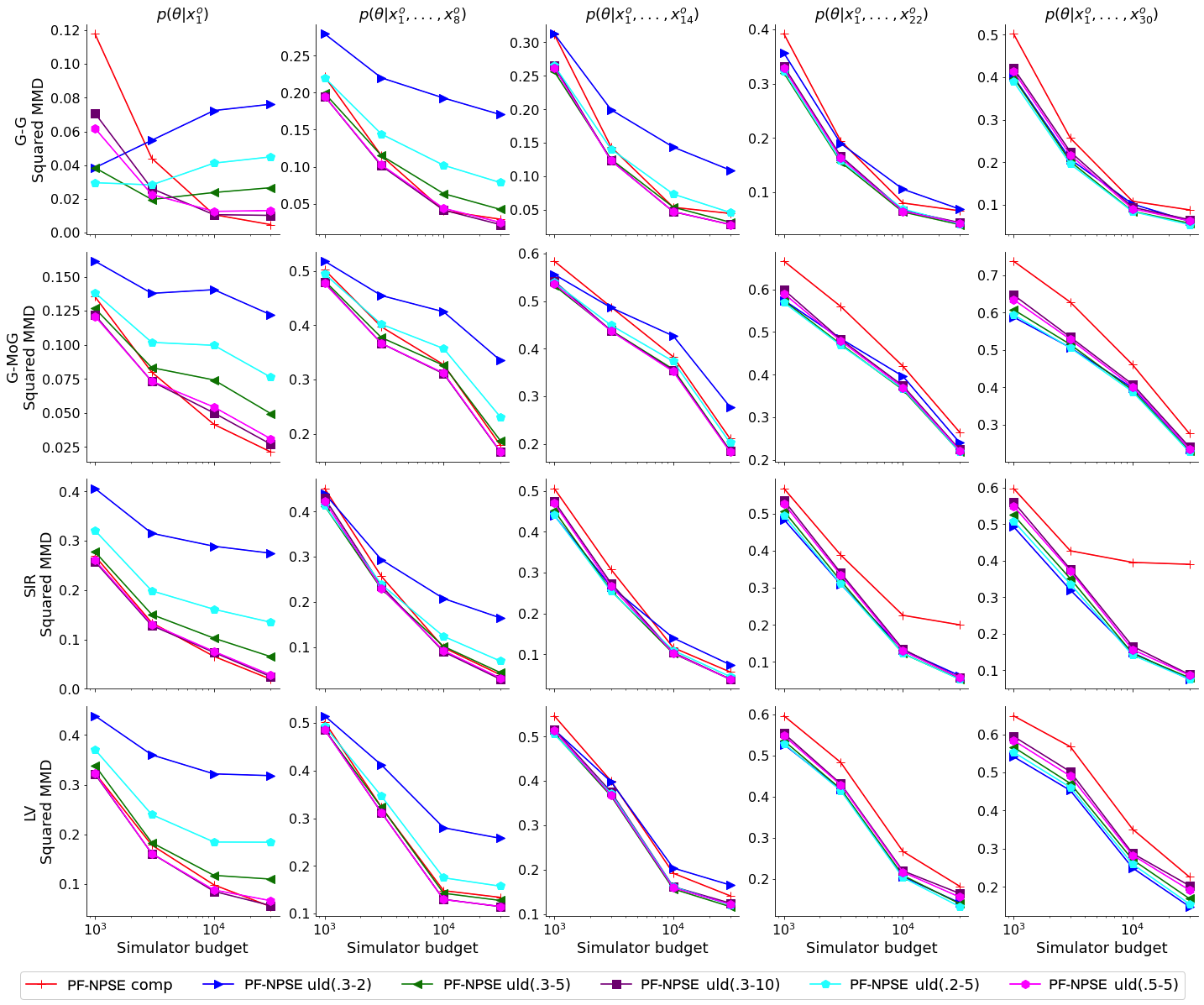}
  
  \caption{Squared MMD (lower is better) achieved by \pfnspe ($m=6$) for different samplers. \textit{comp} indicates sampling using \cref{alg:samplenould} (presented in \cref{app:samplingnould}). \textit{uld(a-L)} indicates sampling using annealed Langevin dynamics (\cref{alg:sampleuld}) with \textit{L} steps per noise level and step-sizes $\delta_t = a \frac{1-\alpha_t}{\sqrt{\alpha_t}}$, where $\alpha_1 = \gamma_1$ and $\alpha_t = \frac{\gamma_t}{\gamma_{t-1}}$ for $t= 2, \hdots, T-1$.}
  \label{fig:ldsampler}
\end{figure*}

\Cref{fig:ldsampler} shows the results obtained using \pfnspe ($m=6$) with different parameters for the annealed Langevin sampler (\cref{alg:sampleuld}), and using the sampling method presented in \cref{alg:samplenould} (\cref{app:samplingnould}), identified with \textit{comp} in the figure's legend. The method's performance is robust to different parameters choices, as long as enough Langevin steps are taken at each noise level. We observe that 5 steps are often enough. Additionally, the results show that the sampling algorithm described in \cref{app:samplingnould} (which does not rely on annealed Langevin dynamics) performs slightly worse than the Langevin sampler.

\clearpage

\end{document}